# SURYA NAMASKAR: REAL-TIME ADVANCED YOGA POSE RECOGNITION AND CORRECTION FOR SMART HEALTHCARE

| Abhishek Sharma | Pranjal Sharma | Darshan Pincha | Prateek Jain |
|---|---|---|---|
| ECE | CCE | CCE | SENSE |
| The LNMIIT, India. | The LNMIIT, India | The LNMIIT, India | VIT AP University, India |

September 6, 2022

## ABSTRACT

Nowadays, yoga has gained worldwide attention because of increasing levels of stress in the modern way of life, and there are many ways or resources to learn yoga. The word yoga means a deep connection between the mind and body. Today there is substantial Medical and scientific evidence to show that the very fundamentals of the activity of our brain, our chemistry even our genetic content can be changed by practicing different systems of yoga. Suryanamaskar, also known as salute to the sun, is a yoga practice that combines eight different forms and 12 asanas(4 asana get repeated) devoted to the Hindu Sun God, Surya. Suryanamaskar offers a number of health benefits such as strengthening muscles and helping to control blood sugar levels. Here the Mediapipe Library is used to analyze Surya namaskar situations. Standing is detected in real time with advanced software, as one performs Surya namaskar in front of the camera. The class divider identifies the form as one of the following: Pranamasana, Hasta Padasana, Hasta Uttanasana, Ashwa - Sanchalanasana, Ashtanga Namaskar, Dandasana, or Bhujangasana and Svanasana. Deep learning-based techniques(CNN) are used to develope this model with model accuracy of 98.68 percent and accuracy score of 0.75 to detect correct yoga (Surya Namaskar ) posture. With this method, the users can practice the desired pose and can check if the pose that the person is doing is correct or not. It will help in doing all the different poses of surya namaskar correctly and increase the effeciency of the yoga practitioner. This paper describes the whole framework which is to be implemented in the model.

## 1 Introduction

As with all exercises, it is very important to do Yoga or Surya Namaskar at a precise positioning as any unusual position does not work and can tend to hurt the person. In the modern world we suggest having a Tutor around while doing yoga.It does not always happen to have a Tutor or join yoga classes these days. So an AI-based program helps in identifying Surya Namaskar conditions properly and provides feedback or suggestions to users. These instructions help users improve their posture to be productive and Correct and not getting hurt by doing the wrong Posture .

So the challenges in this project are significant points should be earned without missing points too; models should work well even where body parts overlap to each other. Suggestions should be given accurately from then on small changes may result in harm. All Surya Namaskar Poses in the database used for this project are developed professionally. Models should accurately distinguish the shape, however they are almost identical to the shape with little difference in them. Automatic training methods for sports Activities it can help players improve their performance and slow down injury risk. Many researchers have developed computer programs to test the effects of exercise . We proposed a 'Surya Namaskar' pose Detection project using accelerated intensity features to make a difference in standing between a student and specialist . Deep Pose has led to off-charge charging deep network-based strategies It retrieves directly from shared links using deep neural network-based regressors. Expect human activities and predicts the location of body parts.However, the approach is difficult to localize. In recent years, there have been activities related to yoga pose discovery and categories. Ways to get key points used by OpenPose, PoseNet , and PifPaf . To find a person's shape, many features will be considered such as environment, human interaction, and diverse clothing . So for the Surya Namaskar Pose Detection The in-depth learning methods which we have used to stand Classes are a multilayer perceptron, Longshort-term memory (LSTM) , a repetitive neural network, and convolutional neural network(CNN) . Limitations on the above functions that features (key points) are not rated and cannot detect a person's pattern of distances of different distances from the camera.

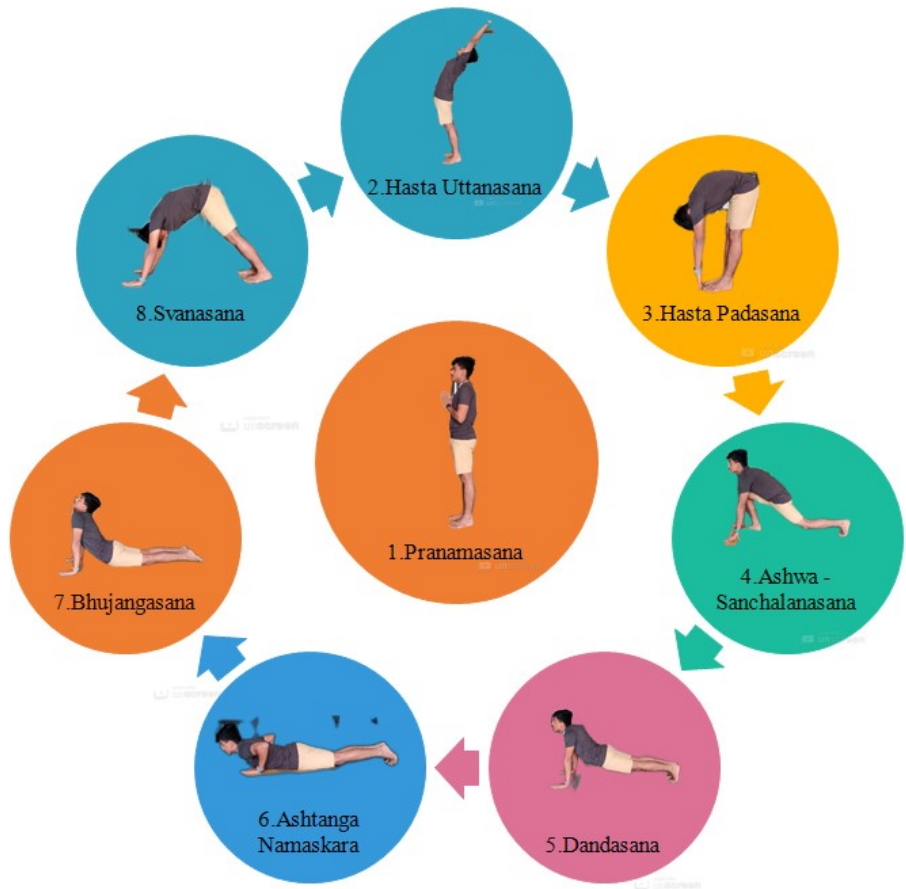

Figure 1: Surya Namaskar Poses

In this project we have estimated different poses in Surya namaskar using pose detection. In this project we have used many no.of frames that is actually predicting what pose is being performed at that point in time,for example we are performing bhujangasana it's actually going to take the entire set of frames for that particular action to go and determine what pose is being performed.The end goal is to produce real time yoga poses in Surya Namaskar pose detection we have done this using python and build it step by step to be able to detect a bunch of different yoga poses and specially Surya Namaskar ,in order to do that we have used few key models so we have used media pipe holistics to extract key points so that had allowed us to extract different key point from diffrent parts of our body. We have used tensor flow and keras and build a model to predict the different yoga poses which are shown on the screen .We have taken Media Pipe holistics and take a trained LSTM model and actually predict the different yoga poses in real time. So first we have collected a bunch of data on all of the different key points .We have collected data from different parts of our body and we have saved those as numpy arrays and then we have trained a deep neural network using LSTM layers and predict the temporal components. So we have predicted the action from a number of frames not just a single frame then we have combined it all together using open CV and predicted it in real time using a webcam.

## 2 Surya Namaskar for Healthcare

Surya kriya or Surya namaskar is an effort to bring our system in syncronicity with rest of our solar system. It is called Surya namkaskar mainly to activate solar plexus, to generate the heat in the solar system to raise samath parana or solar heat in the system. Our spiritual health,your psychological health your spiritual possibility everything is determinde by this.

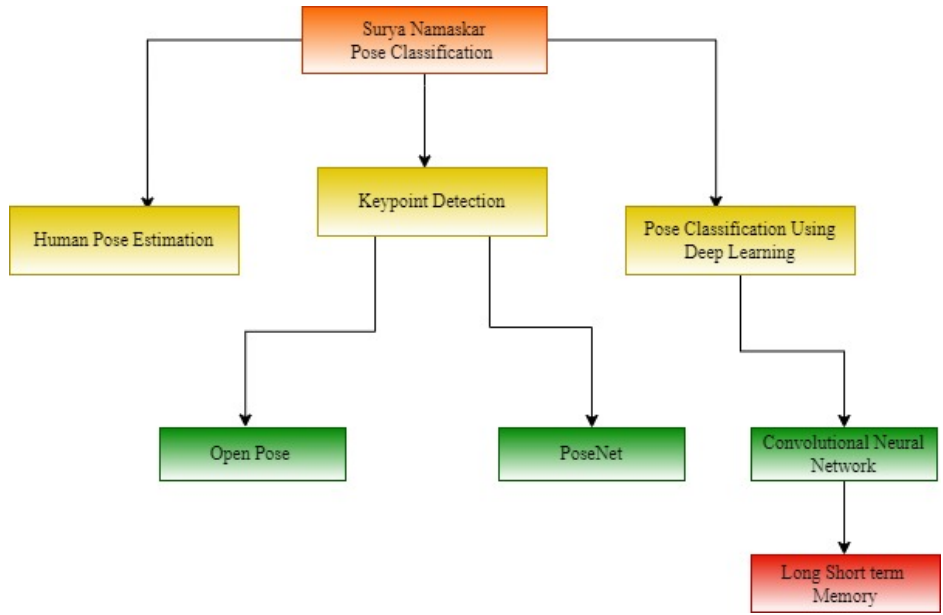

Figure 2: Surya Namaskar Pose Classification

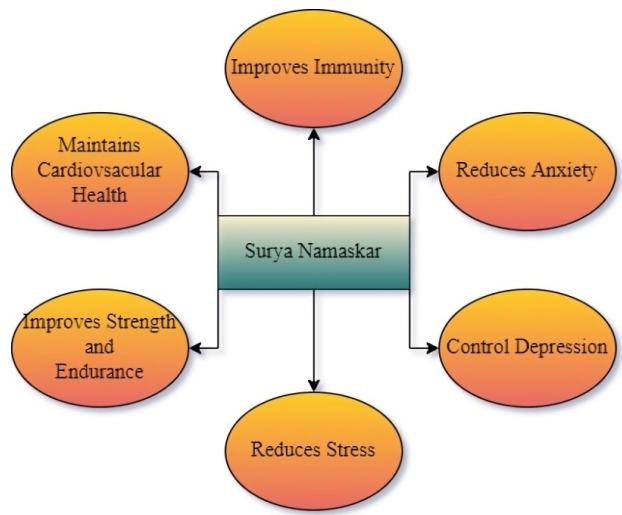

Figure 3: Benefits of Surya Namaskar on regular basis

## 3 Methodology

In order to perform Human Pose Detection using key points there are 11 steps that are followed

1. **Import and install Dependencies:** The first step is to import and install some dependencies, tensor flow 2.8 version ,open-cv, media pipe, sklearn, matplotlib libraries have been used. Open CV is a computer vision library that allows to work with webcams and make it a little bit easier to build field. Open CV is used to access webcam to extract key points and then media pipe holistics are used to actually extract key points. Sklearn is used for evaluation Matrix as well as to leverage a training and testing split and then matplotlib is used that helps to visualize images a little bit easier.

2. **Key points using Media Pipe Holistics:** In second step it is made sure that webcam could be accessed using open CV and then secondary layer is applied where actual detections are made. Media pipe library is used for setting a video capture and then gone to loop through every single frame and actually render that to the screen.

3. **Extract Key points Values:** In Third step key point values were extracted. The key point values were concatenated into numpy array and if the values at approximate time were not available then number zero array was created.

4. **Setup Folder for collection:** In step four folders for array collection were set up, output as a result of going through a data collection of these key points were collected. key points effectively form a Frame values.

5. **Collect keypoint values for Training and Testing:** In step 5 data has been collected , to collect the data, mediapipe loop was started and then specifically snapshots were taken at each point in time. For each one of the actions collection of the actions were taken through the set of frames per video.

6. **Preprocess Data and create labels and features:** In step 6 from sklearn library train test split has been imported that helped in creating a training and testing partition.two category functions were imported from keras that helped with labels.Trained tests split functions is used to partition the data into training partition and in a testing partition.It helped to train one partition and test out on a different segment of data.In this step a label array or label dictionary to represent each one of a diffrent pose was created.

7. **Build and Train LSTM Neural Network:** In 7 step neural network were trained for that tensorflow and keras were used.

8. **Make prediction:** In 8 step predictions were made that were actually passed through a test data set.

9. **Save weights:** Model is saved for further use.

10. **Evaluation using Confusion Matrix and Accuracy:** In 10 step how the model is actually performing was evaluated. For that couple of matrix from sklearn to evaluate the performance of the model has been imported.

11. **Test in real time:** The model was tested in real time in the final step.

## 4 DataSet

The proposed methodology is examined on a data set that is manually created by us. This data set includes 8 Surya Namaskar poses, that is - Pranamasana, Hasta Uttanasana, Hasta Padasana, Dandasana, Ashwa - Sanchalanasana, Ashtanga Namaskara,Bhujangasana and Svanasana.Total videos of the 8 poses were recorded inside a room with the help of camera from a distance of some meters the frame per second rate is 60. To have trained models with good accuracy, The videos were recorded from all angles( front , back and side )to get better results.

## 5 Component of Novelty in the Paper

Till today all the research work that has been done on the yoga pose classification classifies different yoga asanas . This model tests surya namaskar pose in real-time which comprises of 12 steps performed in a specific series of 8 diffrent asanas. This model recognizes each of the 8 asanas. , recognition and grooming for human life. This real-time frame-work is portrayed as a visual yoga teacher with a slight mistake in pose standing.

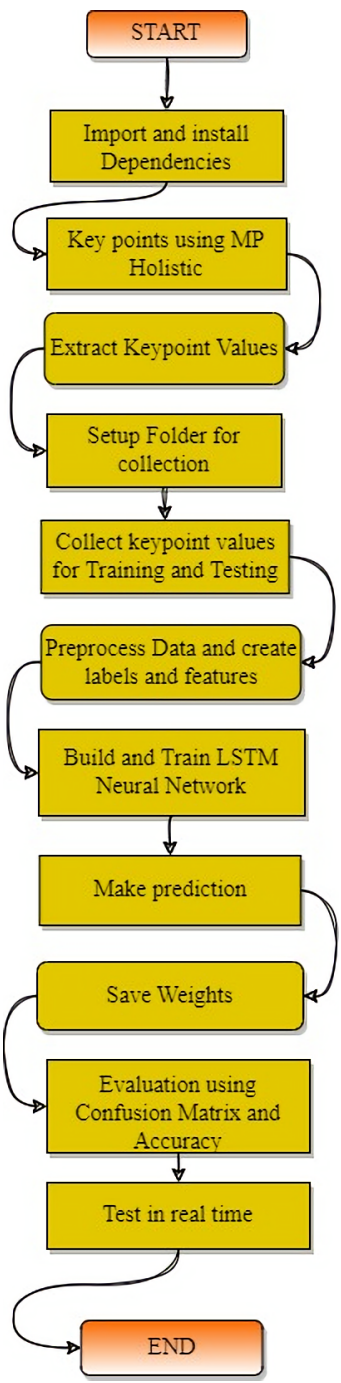

Figure 4: Methodology

# 6 Experimental Analysis and Results

This model predicts the Surya Namaskar poses sequence that is being performed by the user in real-time enviroment and can examine the correctfullness of the poses. The Surya Namaskar pose classification system is evaluated with the help of following criteria shown below as:

1. **Classification Score:** Classification scores usually indicate what model accuracy means. It can be described as the ratio of the number of correct predictions to the total number of input samples. Accuracy is equal to ratio of no. of predictions that are correct to total no. of predictions.

2. **Confusion matrix:** A confusion matrix is one that fully describes the accuracy of the model.

    - you can use confusion matrix to calculate precision, recall, accuracy score, and other metrics. and use that information to calculate precision and recall.
    - Then use precision and recall to get fl-score t will be easier to show you how to calculate these things.
    - you can calculate true negative, true positive, false negative, and false positive numbers,

    Since the diagonal values represent correctly classified samples, we always want the diagonal of the matrix to contain the maximum.

3. **Model accuracy and model loss curve:** Those curves are also referred to as learning curves and are by and large used for fashions that analyze incrementally through the years, as an instance, neural networks. They constitute the assessment on the education and validation information which gives us an idea of the way properly the version is studying and how nicely is it generalizing. The model loss curve represents a minimizing score (loss), which means that a decrease score outcomes in higher model overall performance. Better rating denotes higher performance of the version. a terrific fitting version loss curve is one wherein the education and validation loss lower and attain a factor of stability and feature minimal hole between the very last loss values. then again, the fitted model accuracy curve is a curve this is stabilized with the aid of increasing the schooling and validation accuracy , and there is a minimum gap between the very last accuracy values.

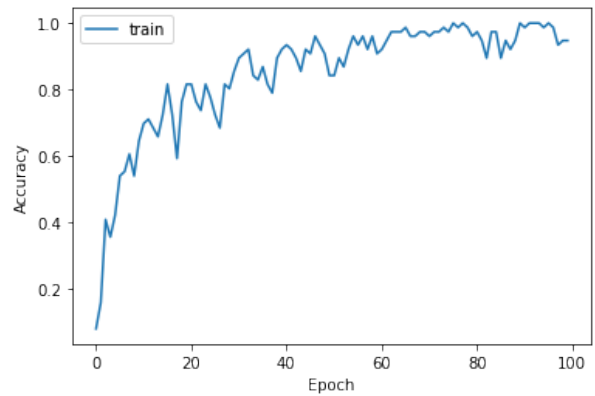

Figure 5: Accuracy vs epoch

Generally as we increases the epochs the Accuracy increases but in some of cases the epochs has increased but the accuracy decreases because of overfitting,exploding gradient and class imbalance problem sometimes the accuracy decreases when increasing the epochs and as we see the Accuracy increases the loss decreases.

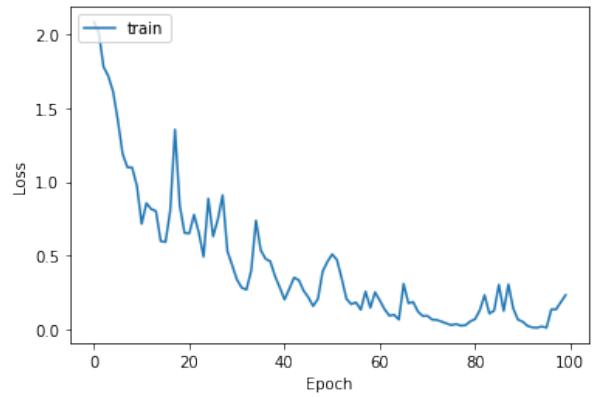

Figure 6: Epoch vs Loss

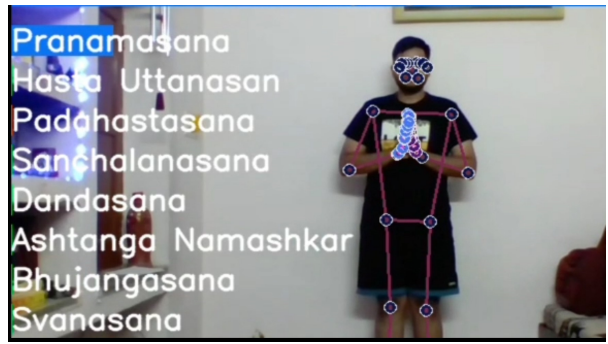

Figure 7: Model accuracy score

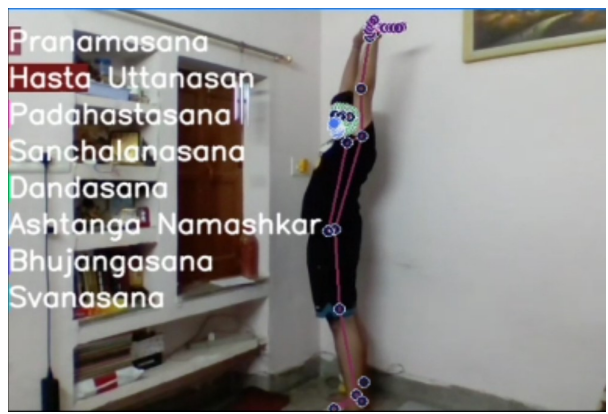

Figure 8: Predicted Output:Pranamasna

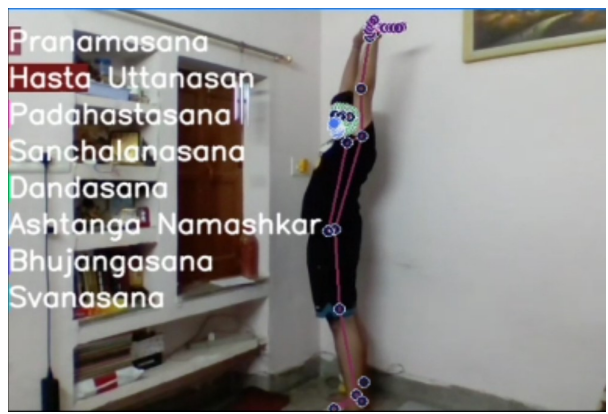

Figure 9: Predicted Output:Hasta Uttanasana

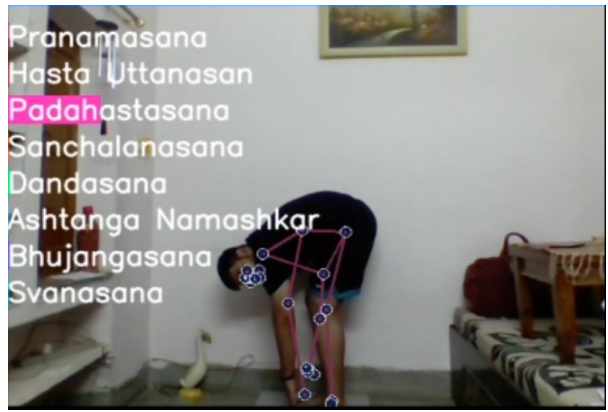

Figure 10: Predicted Output:Padahastasana

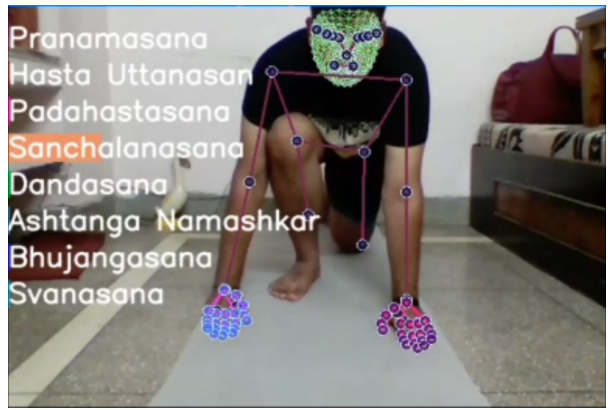

Figure 11: Predicted Output:Sanchalasana

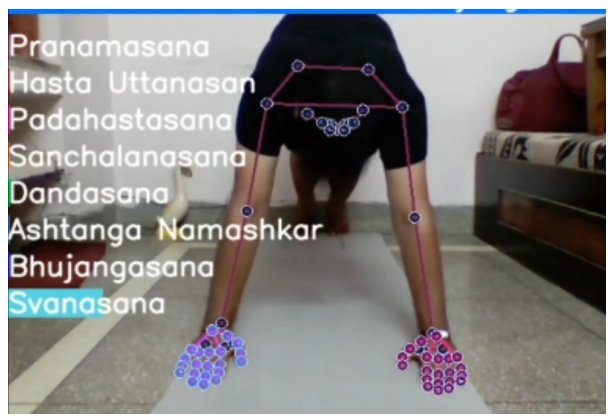

Figure 12: Predicted Output:Svanasana

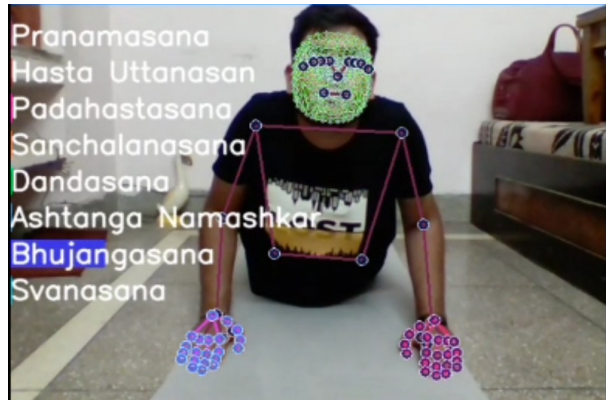

Figure 13: Predicted Output:Bhujangasana

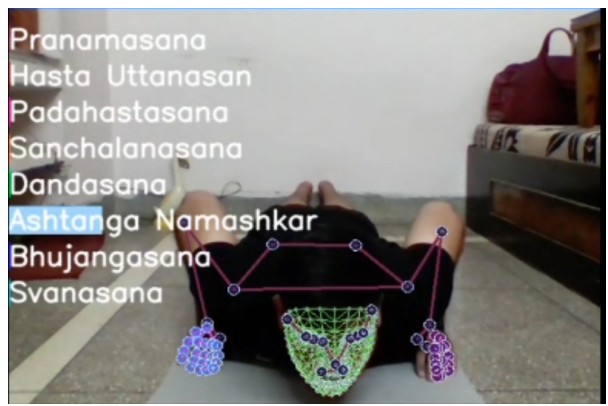

Figure 14: Predicted Output:Ashtanga Namaskar

Table 1: Model Summary Table

| Layer (Type) | Output Shape | Parameters |
| --- | --- | --- |
| lstm_13 (LSTM) | (None, 10, 64) | 442112 |
| lstm_22 (LSTM) | (None, 10, 128) | 98816 |
| lstm_14 (LSTM) | (None, 64) | 49408 |
| dense_12 (Dense) | (None, 64) | 4160 |
| dense_13 (Dense) | (None, 32) | 2080 |
| dense_14 (Dense) | (None, 8) | 264 |
| Total params: 596,840 | Trainable params: 596,840 | Non-trainable params: 0 |

# 7  Conclusion ad Future Development

Estimation of human posture has been studied extensively in recent years. Estimating human posture differs from other computer vision tasks in that it requires localizing and assembling parts of the human body based on an already defined human body structure. Applying postural assessment to physical activities can help prevent injuries and increase the effectiveness of people's workouts. We propose that a yoga self-learning system can not only popularize yoga, but also ensure that it is performed correctly. Deep learning methods are promising due to the extensive research being done in this field. The use of hybrid CNN and LSTM models on OpenPose data is considered very effective and perfectly classifies all eight yoga postures of Surya Namaskar. It is also possible to use native CNNs and SVMs, but SVMs do not work with large datasets. This is less efficient than CNN + LSTM based models. Therefore, this project will help people to perform Surya Namaskar posture accurately and effectively without a mentor.

## About the Authors

**Abhishek Sharma** is an Assistant Professor in Department of ECE, The LNMIIT, Jaipur, India. Contact him at abhisheksharma@lnmiit.ac.in.

**Pranjal Sharma** is currently pursuing undergraduate in Communication and Computer Engineering from The LNMIIT, Jaipur, India. Contact him at 19ucc144@lnmiit.ac.in

**Darshan Pincha** is currently pursuing undergraduate in Communication and Computer Engineering from The LNMIIT, Jaipur, India. Contact him at 19ucc123@lnmiit.ac.in

**Prateek Jain** is currently Assistant Professor in SENSE from VIT AP University, India.  Contact him at prtk.ieju@gmail.com